\documentclass[onecolumn]{article}
\usepackage{amsmath}
\usepackage{graphicx}
\usepackage{hyperref}       
\usepackage{url}            
\usepackage[english]{babel}
\usepackage[nottoc]{tocbibind}
\usepackage[square,numbers]{natbib}
\usepackage{authblk}
\graphicspath{ {./images/} }

\title{\textbf{Re-evaluating the Memory-balanced Pipeline Parallelism: BPipe}}

\author{Mincong Huang\thanks{Contact email:   huangmincong@meituan.com}}
\author{Chao Wang}
\author{Chi Ma}
\author{Yineng Zhang}
\author{Peng Zhang}
\author{Lei Yu}
\affil{Meituan}
\date{}
\begin{document}
\maketitle
\begin{abstract}

Pipeline parallelism is an essential technique in the training of large-scale Transformer models. However, it suffers from imbalanced memory consumption, leading to insufficient memory utilization. The BPipe technique was proposed to address this issue and has proven effective in the GPT-3 model. Nevertheless, our experiments have not yielded similar benefits for LLaMA training. Additionally, BPipe only yields negligible benefits for GPT-3 training when applying flash attention. We analyze the underlying causes of the divergent performance of BPipe on GPT-3 and LLaMA. Furthermore, we introduce a novel method to estimate the performance of BPipe.
\end{abstract}

\section{Introduction}
As Transformer models increase in size, model parallelism becomes essential for efficient training by distributing activations, parameters and optimizer states across devices. Two primary types are tensor parallelism \cite{Shoeybi2019MegatronLMTM} and pipeline parallelism \cite{huang2019gpipe,narayanan2019pipedream,narayanan2021memory,fan2021dapple}. Tensor parallelism divides Attention and FFN layers, whereas sequence parallelism\cite{korthikanti2023reducing} addresses LayerNorm and Dropout layers.
In pipeline parallelism, the \textbf{1F1B}(one-forward and one-backward) strategy\cite{narayanan2021memory,fan2021dapple} is common but suffers from memory imbalance. \textbf{BPipe}\cite{Kim2023BPipeMP} addresses this issue, demonstrating benefits for GPT-3 96B. To confirm this, we implemented BPipe based on Megatron-LM\cite{Megatron-LM}. Our experimental results align with BPipe on GPT-3 96B, but the improvement with flash attention is minor. For LLaMA 65B, BPipe even has a negative impact. These results underscore the varied impact of BPipe across models, urging further exploration and optimization.

\section{Preliminary}
\subsection{Notations}
For the remaining part of this paper, we will use the notations in table \ref{notations}.
\begin{table}
\begin{center}

\begin{tabular}{ c | c  c | c } 
 
 $a$ & number of attention head & $p$ & pipeline parallel size \\
 $b$ & microbatch size & $s$ & sequence length \\
 $h$ & hidden dimension size & $t$ & tensor parallel size \\
 $l$ & number of transformer layers & $v$ & vocabulary size \\
 $B$ & global batch size &  & 
\end{tabular}
\caption{Notations}
\label{notations}
\end{center}
\end{table}
\subsection{BPipe}
\begin{figure}[h]
\centering
\includegraphics[width=1\textwidth]{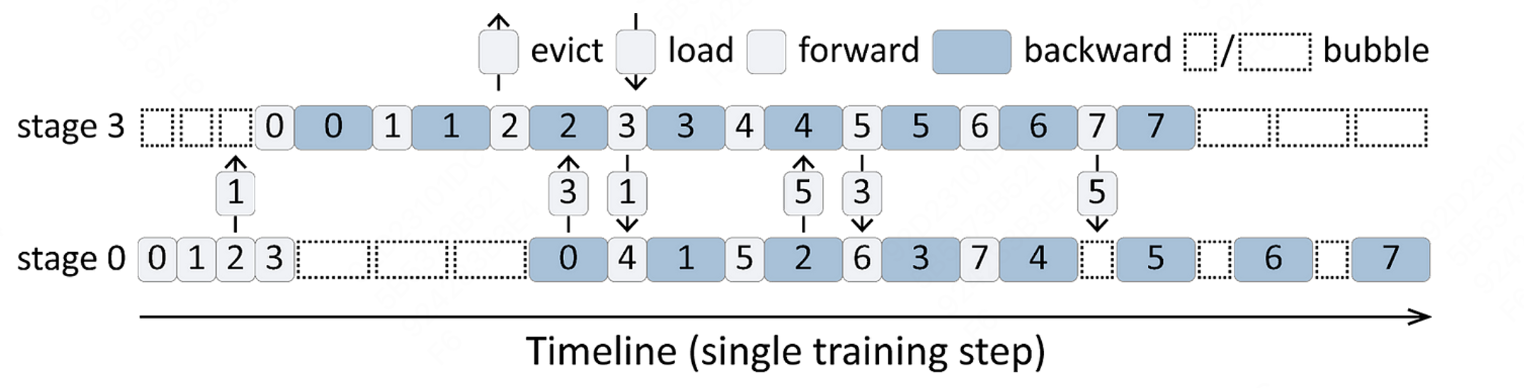}
\caption{An illustration of BPipe within 4-way 1F1B pipeline strategy}
\label{figure bpipe}
\end{figure}

\begin{figure}[h]
\centering
\includegraphics[width=1\textwidth]{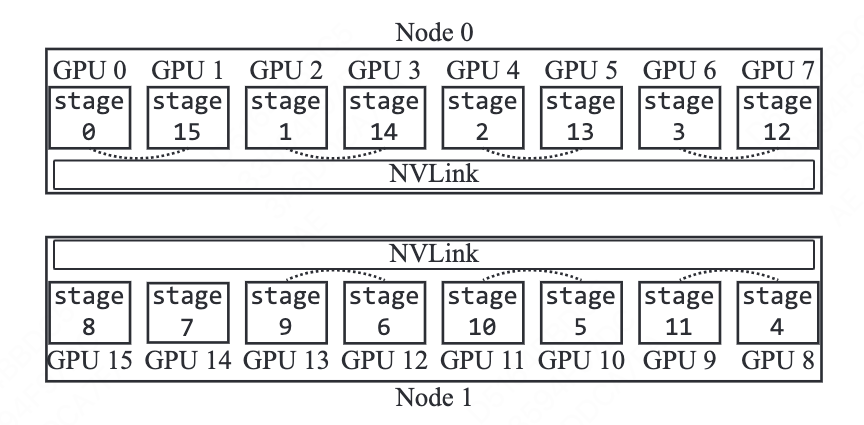}
\caption{An illustration of pair-adjacent
assignment for 16-way pipeline parallelism on two nodes, each
with 8 GPUs}
\label{figure pair adjacent}
\end{figure}

Pipeline parallelism splits the model into multiple stages, facilitating the training of larger models. There are various schedule strategies for pipeline parallelism, like GPipe\cite{huang2019gpipe}, PipeDream\cite{narayanan2019pipedream}, PipeDream-2BW\cite{narayanan2021memory}, DAPPLE\cite{fan2021dapple} and so forth. DAPPLE, a popular choice, is implemented in several prevalent frameworks such as Megatron-LM\cite{Megatron-LM}. This strategy partitions the model parameters, optimizer state, and gradients but does not reduce the activation. When the model is divided into p stages, the stage $x$ of the pipeline model needs to store $p-x$ activations. At stage 0, it needs to store $p$ activations, nearly equal to the activation generated by the all stages once forward propagation. Obviously, this leads to memory imbalance among devices. To solve this problem, BPipe technique was proposed, balancing the activations among devices and ensuring that the number of activations in each device does not exceed $ \lceil \frac{p+2}{2} \rceil$. The basic idea of BPipe is to transfer activation from one device to another. For instance, for the stage $x$($ x < \lfloor \frac{p}{2} \rfloor $) of pipeline parallelism, when the number of activations is about to exceed $ \lceil \frac{p+2}{2} \rceil$, it sends an activation to stage $p-x-1$, and if the activation needed for backward propagation is not in the current device, it loads the activation from the stage $p-x-1$(as figure \ref{figure bpipe} shows). Stage $x$ is referred to as \textbf{Evictor} and stage $p-x-1$ as \textbf{Acceptor}. The additional communication can overlap with forward or backward computation. If the evictor-acceptor pair is within the same node of the cluster, the communication time is typically less than the forward or backward propagation due to NVLink. However, for larger models, the evictor-acceptor pair may not be within the same node. In such cases, the layout of the evictor-acceptor pairs should be arranged to ensure they are within the same node, as figure \ref{figure pair adjacent} shows.

\section{Experiment and Analysis}
\subsection{Experiments}
We conducted evaluations on LLaMA 65B model and GPT-3 96B model (the configuration of models are detailed in table \ref{table:config of models}). Our evaluations were conducted on a cluster of 4 nodes, each of which is equipped with 8 NVIDIA 80 GiB A100 GPUs connected over NVLink. The parallelism strategies  were set at $t=4$ and $p=8$, and global batch size was 128. The framework we used was Megatron-LM\cite{Megatron-LM}, and enabled sequence parallelism technique. We trained the models in mixed precision\cite{micikevicius2017mixed} with the Adam\cite{kingma2014adam} optimizer.
\begin{table}
\begin{center}

\begin{tabular}{ c | c  c  c  c  c } 
 \hline
 Model & $h$ & $a$ & $s$ & $l$ & $B$ \\ [0.5ex] 
 \hline
 LLaMA 65B & 8192 & 64 & 2048 & 80 & 128 \\
 GPT-3 96B & 9984 & 104 & 2048 & 80 & 128 \\ [1ex]
 \hline
\end{tabular}
\caption{Configuration of models}
\label{table:config of models}
\end{center}
\end{table}

As the same as BPipe paper, we use model FLOPS utilization (MFU)\cite{chowdhery2022palm} as the evaluation
metric, which is a ratio of the observed throughput to the hard-
ware maximum throughput. For the observed throughput calculation, we refer to related method\cite{korthikanti2023reducing,narayanan2021efficient}, which mainly considers the contribution of matrix multiplication to FLOPs, and gives the FLOPs calculation formula of GPT-3(equation \ref{model FLOPs}). 
\begin{equation}
72bslh^2(1+\frac{s}{6h}+\frac{v}{16lh})
\label{model FLOPs}
\end{equation}
For the matrix multiplication, the main difference between LLaMA and GPT-3 is in the FFN. The FLOPs of the FFN of GPT-3 Model can be calculated with the formula $ 16bsh^2 $. There are three matrix multiplication operations in the FFN of LLaMA model, each of which either increases the hidden size to $ \frac{8}{3}h $ or reduces it back to $h$. So we get $ 3\times 2\times \frac{ 8}{3} bsh^2 = 16bsh^2 $ floating-point operations(factor of 2 needed to account for multiplies and adds), which is the same as GPT-3 model, and it means we can use the same formalu as GPT-3 model to caculate the FLOPs of LLaMA model.

In BPipe paper, it mentions using activation checkpointing\cite{chen2016training} technique to reduce the memory for the larger micro batch size. In our experiments, we only recompute the attention layer\cite{korthikanti2023reducing}. In addition, we also use the flash attention\cite{dao2022flashattention,dao2023flashattention} to substitute the recomputing of attention, as flash attention also does not store the intermediate activation in the attention layer and is more efficient. The results of our experiments are showed in table \ref{table:entire model}.
\begin{table}
\begin{center}

\begin{tabular}{ c | c | c | c  c | c } 
 \hline
 Model & ID & $b$ & BPipe & attention method & MFU [\%] \\ [0.5ex] 
 \hline\hline
 LLaMA 65B & (1) & 1 & No & none & 45.3 \\
           & (2) & 2 & No & recompute & 46.0 \\
           & (3) & 4 & Yes & recompute & 42.7 \\
           & (4) & 1 & No & flash attn 2 & 47.8 \\
           & (5) & 2 & No & flash attn 2 & 49.2 \\
           & (6) & 4 & Yes & flash attn 2 & 44.0 \\
 \hline
 GPT-3 96B & (7) & 1 & No & recompute & 34.0 \\
           & (8) & 2 & Yes & recompute & 45.8 \\
           & (9) & 1 & No & flash attn 2 & 52.0 \\
           & (10) & 2 & Yes & flash attn 2 & 51.7 \\ [1ex]
 \hline
\end{tabular}
\caption{Experiments of the models. For attention, we use different implementations. None means we used the original attention. Recompute means we use activation checkpointing technique on the attention. Flash attn 2 means we use the flash attention 2\cite{FlashAttention} to substitute the original implementation.} 
\label{table:entire model}
\end{center}
\end{table}

\subsection{Analysis}
The MFU of experiment (7) and (8) is lower than that reported in the BPipe paper, but similar speedup is achieved. We attribute this discrepancy to the difference between our environments. This doesn't impact our analysis as we primarily focus on speedup rather than MFU values. However, in the experiments (9) and (10), we get a negative result with the flash attention. Additionally, we note that in the BPipe paper, it reports MFU of 48.78\% with $t=2$, $p=16$, $b=1$ and attention recomputing, which is close to its best result and is much higher than our result in experiment (7). This contradicts our experience, as we typically observe better performance when transitioning from $t=2$ to $t=4$.

We conducted profiling with the configurations of Experiments (7) and (8), and got the reason for the observed results. In Experiment (7), some inefficient GPU kernels run for the scale and softmax operation in the Attention layer. It initially converts fp16/bf16 to fp32, applies scale and softmax, and then converts the fp32 back to fp16/bf16. These memory-bound operations would be more efficient if fused into a single kernel. Such fusion is implemented in the Megatron-LM framework, and in Experiment (8), it runs with this fused kernel, resulting in improved performance. In Experiments (9) and (10), we use flash attention, eliminating kernel differences and leading to a negative result.

We noticed that the authors of BPipe paper conducted additional experiments using flash attention, yielding disparate results. They claim that with flash attention, they get 1.32x speedup. It is perplexing, because with the different kernel implementations, they get 1.36x speedup, which is close to the result with flash attention.

\section{Performance Estimation}
Although we haven't seen significant improvements with BPipe, it remains an effective technique for memory reduction with minimal overhead. Actually, the implementation of BPipe is not straightforward, so we introduce a method to estimate its potential benefits. As we primarily use the reduced device memory to increase the micro batch size, we have conducted the analysis towards this aspect. For convenience, we define more notations in table \ref{table:notations names 2}. We disregard the communication time of pipeline parallelism and the overhead of the optimizer as they are typically negligible in training. We also temporarily ignore the overhead introduced by the BPipe technique. For pipeline parallelism, there are bubbles and the time spent in bubbles is approximately $(p-1)\times T(b)$, hence the MFU of the model can be defined as in equation \ref{MFU simple}.
\begin{table}
\begin{center}

\begin{tabular}{ c | c } 
 
 $T(b)$ &  the total time of a forward propagation and \\ & a backward propagation in a single pipeline stage \\
 $P$ & the theoretical maximum FLOPS of the device \\
 $F$ & the FLOPs of the model \\
 $F_{stage}$ & the FLOPs of a single stage in pipeline model \\
\end{tabular}
\caption{Notations}
\label{table:notations names 2}
\end{center}
\end{table}

\begin{equation}
\label{MFU simple}
MFU(b)=\frac{1}{P}\frac{F}{(\frac{B}{b}+p-1)\times T(b)}
\end{equation}

The value of $T$ is directly related to the MFU that a single stage of the model can achieve, and MFU can directly indicate performance. Therefore, we introduce the MFU of a single stage as  $MFU_{stage}(b)=\frac{1}{P}\frac{b\times F_{stage}}{B\times T(b)}$. Finally, we can determine the MFU of the model as equation \ref{MFU final}.
\begin{equation}
\label{MFU final}
\begin{split}
    MFU(b) &= \frac{1}{P}\frac{F}{(\frac{B}{b}+p-1)\times \frac{b\times F_{stage}}{P \times B \times MFU_{stage}(b)}}\\
    &= \frac{F \times {MFU_{stage}(b)} }{(1 +\frac{b}{B}\times (p-1))\times F_{stage}}
\end{split}
\end{equation}
The relationship between the performance of the model and the performance of a single stage can be obtained as equation \ref{MFU ratio}.
\begin{equation}
\label{MFU ratio}
    \frac{MFU(x)}{MFU(y)} = \frac{B+y \times (p-1)}{B+x\times (p-1)} \frac{MFU_{stage}(x)}{MFU_{stage}(y)}
\end{equation}

\begin{table}
\begin{center}
\begin{tabular}{ c | c | c |  c | c } 
 \hline
 Model & ID & $b$ &  attention method & MFU of a single stage [\%] \\ [0.5ex] 
 \hline\hline
 LLaMA 65B & (1) & 1 &  none & 51.1 \\
           & (2) & 2 & recompute & 54.5 \\
           & (3) & 4 & recompute & 57.6 \\
           & (4) & 1 & flash attn 2 & 53.6 \\
           & (5) & 2 & flash attn 2 & 58.6 \\
           & (6) & 4 & flash attn 2 & 61.9 \\
 \hline
 GPT-3 96B & (7) & 1 & recompute & 37.8 \\
           & (8) & 2 & recompute & 55.2 \\
           & (9) & 1 & flash attn 2 & 57.7 \\
           & (10) & 2 & flash attn 2 & 62.4 \\ [1ex]
 \hline
\end{tabular}
\caption{Experiments of single stage of model}
\label{table:single stage}
\end{center}
\end{table}
For the experiment (7) and (8), as the micro batch size increases from 1 to 2, the MFU of a single stage increases from 37.8\% to 55.2\%, resulting in 1.46x speedup. We can calculate the expected speedup ratio of the entire model to be $ \frac{128+1\times(8-1)}{128+2\times (8-1)}\times1.46 \approx 1.39 $. The MFU of the entire model increases from 34.0\% to 45.8\% and the speedup ratio is 1.35, which closely aligns with 1.39. We believe the difference primarily stems from the overhead of BPipe technique which is ignored in our analysis. We observe similar results in other experiments. 

Apparently, this estimation method isn't exclusive to the BPipe technique. It provides an approximate upper bound of the speedup following an increase in micro batch size in pipeline parallelism, indicating whether these is a need to apply further techniques to increase the micro batch size. 
\section{Conclusion}
The BPipe technique effectively balances memory in pipeline parallelism, thereby facilitating a potential increase in the micro batch size. However, performance is influenced not only by the micro batch size but also by the model, parallelism configuration, among other factors. Prior to applying the BPipe technique, we can evaluate a small part of the model with fewer resources to estimate the entire model's performance following an increase in the micro batch size.

\bibliographystyle{unsrt}
\bibliography{ref}

\end{document}